\documentclass{article}

\usepackage{arxiv}

\usepackage[utf8]{inputenc} 
\usepackage[T1]{fontenc}    
\usepackage{hyperref}       
\usepackage{url}            
\usepackage{booktabs}       
\usepackage{amsfonts}       
\usepackage{nicefrac}       
\usepackage{microtype}      

\usepackage{lipsum}         
\usepackage{graphicx}
\usepackage{natbib}
\usepackage{doi}

\usepackage{amsmath,amssymb}
\usepackage{cleveref}       

\usepackage[table]{xcolor}   
\usepackage{colortbl}        

\usepackage{subcaption}

\title{Beyond More Context: Retrieval Diversity Boosts Multi-Turn Intent Understanding}

\author{ \href{https://orcid.org/0009-0009-7144-4825}{\includegraphics[scale=0.06]{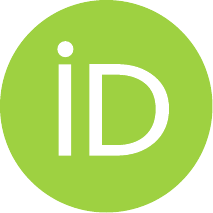}\hspace{1mm}Zhiming Lin} \\
	Independent Researcher\\
	\texttt{nklinzhiming@gmail.com} \\
}

\renewcommand{\shorttitle}{Beyond More Context: Retrieval Diversity Boosts Multi-Turn Intent Understanding}

\hypersetup{
pdftitle={Beyond More Context: Retrieval Diversity Boosts Multi-Turn Intent Understanding},
pdfsubject={q-bio.NC, q-bio.QM},
pdfauthor={David S.~Hippocampus, Elias D.~Striatum},
pdfkeywords={First keyword, Second keyword, More},
}

\begin{document}
\maketitle

\begin{abstract}
Multi–turn intent understanding is central to task–oriented chatbots, yet real deployments face tight token budgets and noisy contexts, and most retrieval pipelines emphasize relevance while overlooking set–level diversity and confounds such as “more context” or exemplar order. 
We ask whether retrieval diversity, rather than longer prompts, systematically improves LLM intent understanding under fixed budgets. 
We present a diversity–aware retrieval framework that selects in–context exemplars to balance intent coverage and linguistic variety, and integrates this selection with standard LLM decoders; the evaluation enforces budget–matched prompts and randomized positions, and includes sensitivity analyses over exemplar count, diversity strength, and backbone size. 
On MultiWOZ~2.4 and SGD, the approach achieves strong gains in Joint Goal Accuracy (JGA) under equal token budgets, surpassing strong LLM/DST baselines, with consistent improvements across $K{=}4\!\sim\!7$ and moderate latency. 
Overall, the study isolates and validates the impact of content diversity in retrieval and offers a simple, deployable selection principle for building accurate, budget–constrained multi–turn intent systems.
\end{abstract}


\section{Introduction}
Task–oriented chatbots have become a core interface for customer service, commerce, and support, where rapid and accurate intent understanding shortens wait time and lowers operational cost\citep{liu2023can,yeh2022guide,cai2022task,xiao2024confusion}. In real deployments, interactions naturally evolve into multi–turn conversations: the same surface form can imply different intents depending on context, while production systems maintain hundreds or thousands of intents, far beyond the label cardinalities of typical emotion or dialogue–act tasks\citep{zhou2018multi,gu2020speaker,kao2019model,fan2022building,tao2019multi,yao2023ndc}. Large language models (LLMs)\citep{naveed2025comprehensive,zhao2023survey,tong2025does} and retrieval–augmented paradigms\citep{gao2023retrieval,chen2025framework} promise to alleviate data and context limitations, yet multi–turn intent understanding remains challenging under tight latency and token budgets.

Despite rapid advances in task-oriented dialogue modeling and intent understanding—spanning classic slot-filling pipelines, neural encoder–decoder architectures for \emph{Dialogue State Tracking (DST)} (e.g., TRADE), description-driven formulations (e.g., D3ST), and, more recently, in-context learning with large language models (e.g., IC-DST, SM2, LDST)~\citep{wu2019transferable,zhao2022description,hu2022context,chen2023stabilized,feng2023towards,zhang2025enhancing}—the selection of contextual evidence still typically prioritizes relevance or relies on heuristic diversification. Prior DST–RAG systems and ICL selectors seldom formalize the set-level trade-off between covering many plausible intents and avoiding textual redundancy under a fixed prompt budget~\citep{venkateswaran2023district,king2023diverse,lewis2020retrieval}. Classic IR diversifiers (e.g., MMR/FPS) improve novelty or aspect spread~\citep{carbonell1998use}, but are agnostic to intent-label coverage and rarely control confounds specific to LLM prompting (token length, exemplar order) that can inflate gains without improving genuine understanding~\citep{liu2022makesgood,lu2022fantastically}. In short, the field lacks a principled, label-aware diversification scheme with fairness controls that isolates whether content diversity—rather than “more tokens” or “better positions”—is the driver of multi-turn intent gains.

This paper asks: \emph{Can diversity in retrieved content systematically improve LLM intent understanding in multi–turn dialogue under fixed budgets, beyond relevance–only retrieval?} We address this by introducing a high–level, label–aware diversification principle that balances intent coverage and linguistic variety for the exemplars supplied to an LLM, together with evaluation protocols that control for token budget and position bias.

Our contributions can be summarized as follows:
(i) \textbf{Insight \#1 (Beyond “more context”)}.
Under matched token budgets and randomized exemplar positions, we show that content diversity—the joint balance of label coverage and linguistic variety—rather than longer prompts or favorable ordering, is the main driver of multi–turn intent gains over relevance–only retrieval and heuristic diversifiers.
(ii) \textbf{Insight \#2 (Set–level trade–off).} We recast exemplar selection as a label-aware balance between covering plausible intents and avoiding textual redundancy, clarifying when and why diversity helps in multi-turn settings and offering practical guidance for choosing exemplar count and diversity strength under fixed budgets.
(iii) \textbf{Performance.} On MultiWOZ and SGD, our diversity–aware retrieval yields consistent, budget-controlled improvements over strong DST/LLM baselines, and the gains persist across backbone sizes and training regimes (zero-/few-shot SFT/RL) while satisfying practical latency constraints.

\section{Related Work}
\subsection{Multi–turn DST: from encoders to LLM/ICL.}
Neural dialogue state tracking(DST)\citep{williams2016dialog} progressed from pointer/copy generators (e.g., TRADE) that share parameters across slots and domains~\citep{wu2019transferable} to description–driven DST (D3ST) that conditions on natural–language slot descriptions for improved transfer~\citep{zhao2022description}. With large LMs, in-context DST emerged: IC-DST retrieves examples for few-shot prompting~\citep{hu2022context}, SM2 stabilizes ICL via prompt/meta learning~\citep{chen2023stabilized}, LDST shows instruction-tuned, open-weight LLMs can approach or match closed models~\citep{feng2023towards}. These works typically treat retrieval as relevance-only or apply heuristics; set-level diversity under token budgets and order/position confounds are rarely analyzed\citep{tong2025rainbow}. We target multi-turn intent with an explicit set objective that blends label coverage and linguistic variety, and we evaluate under matched token budgets with shuffle/position controls.

\subsection{Retrieval–augmented DST.}
Retrieval-Augmented Generation (RAG) introduced non-parametric memory to knowledge-intensive NLP~\citep{lewis2020retrieval}. For DST, DiSTRICT performs retriever-driven in-context tuning~\citep{venkateswaran2023district}, and RefPyDST advocates diverse retrieval and code-style states~\citep{king2023diverse}. Beyond DST, recent ICL selection studies learn or score demonstrations by similarity, influence, or coverage~\citep{rubin2022learningretrieve,liu2022makesgood,min2022rethinking,jiabao2025revibranch}. Prior DST-RAG work seldom formalizes the coverage–redundancy trade-off at the set level, many ICL selectors emphasize relevance or per-pair novelty but do not enforce label coverage. We cast selection as a constrained set optimization with $R(S)=\alpha G(S)+(1-\alpha)D(S)$, where $G$ (Gini over labels) promotes intent coverage and $D$ penalizes average similarity. We show consistent gains over Top-$K$/MMR/FPS at equal token budgets and across $K$.

\begin{figure*}[htb]
	\centering
	\includegraphics[width=0.95\linewidth]{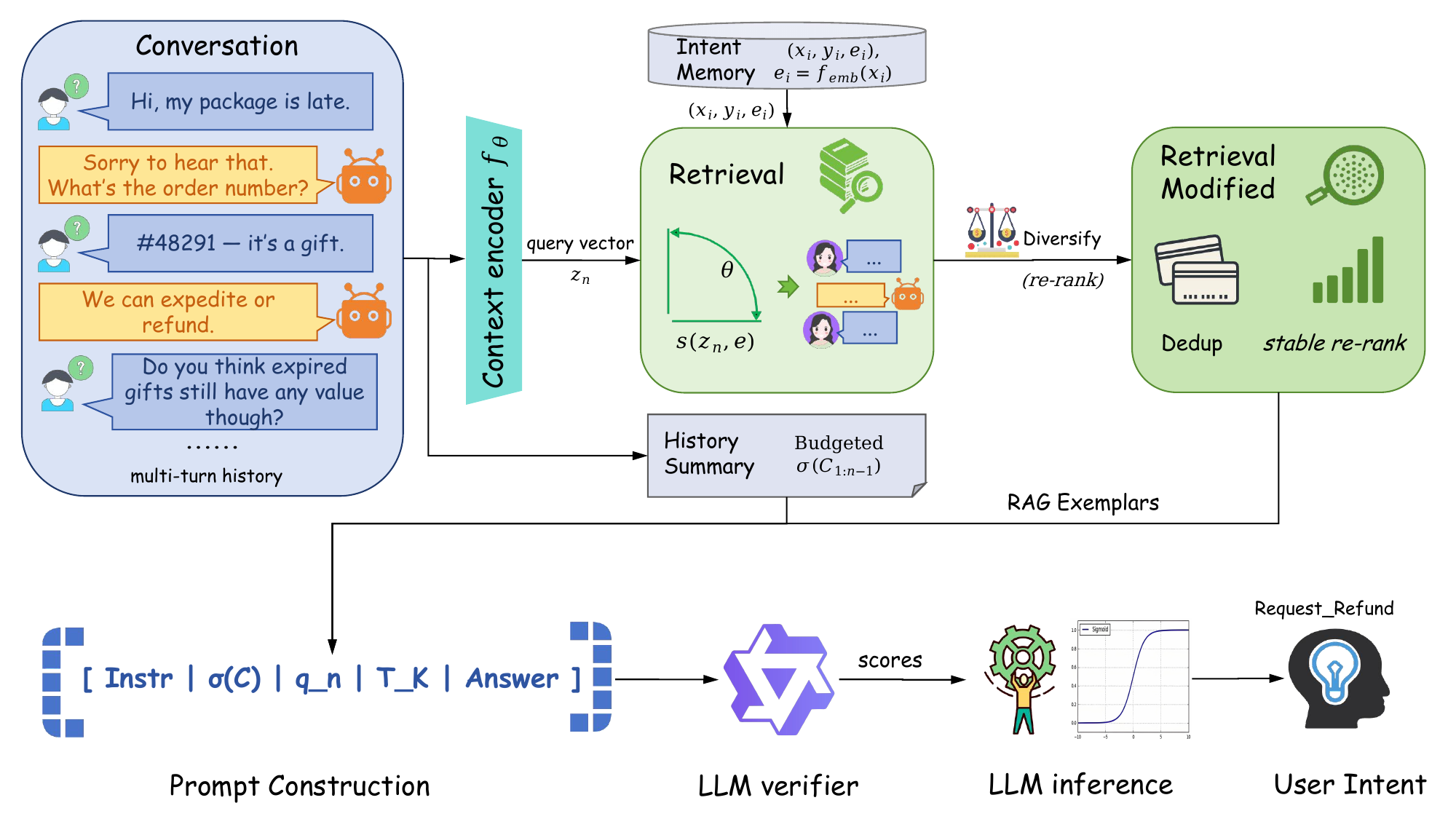}
	\caption{\textbf{LDRA pipeline for multi-turn intent classification.} (Left) An illustrative multi-turn conversation provides history \(C_{1:n-1}\) and the current utterance \(q_n\). The context encoder \(f_\theta\) yields a context-aware query vector \(z_n\). (Top-middle) A hybrid retriever queries the intent memory \(\mathcal{M}=\{(x_i,y_i,e_i)\}\) to form a candidate pool \(S_L\) using \(\mathrm{Rel}(i\!\mid\!z_n)=\lambda_{\mathrm{vec}}\,s(z_n,e_i)+(1-\lambda_{\mathrm{vec}})\,\mathrm{BM25}(q_n,x_i)\). (Middle) A diversity engine re-ranks candidates by maximizing \(R(S)=\alpha\,G(S)+(1-\alpha)\,D(S)\) with \(G(S)=1-\sum_k p_k^2\) (label coverage) and \(D(S)=1-\bar{s}(S)\) (linguistic variety), under constraints \(s(z_n,e_i)\!\ge\!\tau\) and \(n_k\!\le\!U\), producing a diverse set \(T_K\) (“Retrieval Modified”: dedup + stable re-rank). (Bottom) The dialogue history is summarized into \(\sigma(C_{1:n-1})\); together with \(T_K\) it forms the structured prompt \(\Pi=[\text{Instr};\,\sigma(C);\,q_n; \,T_K;\,\text{Answer}]\). The LLM verifier \(\Phi\) outputs calibrated scores \(S_\Phi(y\!\mid\!\Pi)\), and the inference step predicts \(\hat{y}=\arg\max_y S_\Phi(y\!\mid\!\Pi)\) (example intent: \texttt{Request\_Refund}).}
	\label{fig:framework}
\end{figure*}

\subsection{Diversity-aware selection in IR.}
Maximal Marginal Relevance (MMR) balances query relevance with novelty to reduce redundancy~\citep{carbonell1998use}. Farthest-first selection (the $k$-center 2-approximation of Gonzalez) is a classical way to spread selections in metric space~\citep{gonzalez1985clustering}. 
MMR/FPS capture pairwise novelty but are agnostic to label coverage and to the set-wise interaction between coverage and redundancy that matters for intent classification with many labels. 
By mixing $G$ and $D$ and enforcing per-label caps and relevance thresholds, our LDRA explicitly targets semantic coverage of intents while controlling redundancy, which we verify empirically and analyze through sensitivity studies.

\section{Method: LDRA (Linguistic-Diversity Retrieval-Augmentation)}

We propose \textbf{LDRA}, a retrieval-augmented framework that improves multi-turn intent classification by explicitly optimizing the diversity of retrieved supervision signals (intent-labeled exemplars) while preserving query relevance. Given a dialogue context $C_{1:n-1}=\{(q_t,r_t)\}_{t=1}^{n-1}$ and the current user query $q_n$, LDRA (i) encodes a context-aware query vector, (ii) retrieves a relevance-filtered candidate pool, (iii) re-ranks it via a label-diversity and text-diversity objective, and (iv) composes a structured instruction prompt for an LLM-based decoder to predict the final intent $\hat{y}$. The core novelty is an optimization-guided re-ranking that balances coverage of intent categories (via a Gini-style label diversity) and linguistic variety (via average dissimilarity of exemplar texts). Figure \ref{fig:framework} shows the computational process of LDRA.

\subsection{Problem Setting and Notation}
We are given a multi-turn dialogue $C_{1:n-1}=\{(q_t,r_t)\}_{t=1}^{n-1}$ and current utterance $q_n$. The goal is to predict the intent $y^\star \in \mathcal{Y}$ of $q_n$ by leveraging $C_{1:n-1}$ and a supervision memory $\mathcal{M}=\{(x_i,y_i,e_i)\}_{i=1}^N$, where $x_i$ is a labeled exemplar (short user text), $y_i \in \mathcal{Y}$ its intent, and $e_i=f_{\text{emb}}(x_i)\in\mathbb{R}^d$ a precomputed embedding (e.g., from a small fine-tuned encoder). We denote cosine similarity by
\begin{equation}
s(u,v) \triangleq \frac{\langle u,v\rangle}{\|u\|_2\|v\|_2}\in[-1,1],\quad
\bar{s}(S)\triangleq \tfrac{2}{|S|(|S|-1)}\sum_{i<j}s(e_i,e_j)\;\text{for a set }S.
\end{equation}
Here, $s(\cdot,\cdot)$ compares normalized vectors; $\bar{s}(S)$ is the average pairwise similarity within $S$.

\subsection{Context-Aware Query Encoding}
The same surface form may map to different intents depending on conversational state; thus, $q_n$ must be encoded with respect to its history to disambiguate intent.
We compute a context-aware query $z_n\in\mathbb{R}^d$ by cross-attending $q_n$ to history:
\begin{align}
h_t &= f_{\text{emb}}(q_t) \in \mathbb{R}^d,\quad g_t=f_{\text{emb}}(r_t)\in \mathbb{R}^d, \\
\beta_t &\propto \exp\!\Big(\tfrac{(W_q f_{\text{emb}}(q_n))^\top (W_k h_t)}{\sqrt{d}} + \rho\,\kappa(t,n)\Big), \\
\gamma_t &\propto \exp\!\Big(\tfrac{(W_q f_{\text{emb}}(q_n))^\top (W_k g_t)}{\sqrt{d}} + \rho\,\kappa(t,n)\Big), \\
z_n &= \text{LN}\Big( f_{\text{emb}}(q_n) + W_v \sum_{t=1}^{n-1} \beta_t h_t + W_v' \sum_{t=1}^{n-1} \gamma_t g_t \Big).
\end{align}
where $W_q,W_k,W_v,W_v'\in\mathbb{R}^{d\times d}$ are learnable; $\kappa(t,n)=-\lambda(n-t)$ is a recency kernel ($\lambda\ge 0$); $\rho$ weights recency; LN is layer normalization.
$h_t,g_t$ are embeddings of past user/agent turns; $\beta_t,\gamma_t$ are attention weights; $z_n$ blends current turn with history.

\subsection{Similarity Retrieval}
LLM context is limited\citep{yao2024swift}; we must first filter a large memory to a small, relevant candidate pool.
We retrieve a pool $S_L=\{(x_i,y_i,e_i)\}_{i=1}^{L}$ of the top-$L$ items by a hybrid score
\begin{equation}
\label{eq:hybrid}
\operatorname{Rel}(i \mid z_n) \;=\; \lambda_{\text{vec}}\, s(z_n,e_i) \;+\; (1-\lambda_{\text{vec}})\,\operatorname{BM25}(q_n,x_i),
\end{equation}
where $\lambda_{\text{vec}}\in[0,1]$. We use ANN for vector search and optionally BM25 for lexical matching.
$S_L$ is the relevance-filtered set; $L$ is a small pool size before diversification.

\subsection{Diversity-Aware Re-ranking}
Pure similarity retrieval tends to be redundant (near-duplicates), missing alternative phrasings and \emph{distinct intents} that are still plausible in context. We promote \emph{coverage} over intent labels and \emph{linguistic variety} over exemplars to improve intent disambiguation.

\paragraph{Label diversity.}
For a subset $S\subseteq S_L$ with label counts $n_k=|\{i\in S: y_i=k\}|$ and proportions $p_k=\frac{n_k}{|S|}$, define the (Gini-style) label diversity
\begin{equation}
G(S) \;=\; 1 - \sum_{k\in\mathcal{Y}} p_k^2.
\end{equation}

$G(S)\in[0,1)$ increases when labels are more evenly covered; it is $0$ when all items share one label.

\paragraph{Text diversity.}
Let $\bar{s}(S)$ be the mean pairwise cosine similarity within $S$. Define text diversity as the complementary dissimilarity
\begin{equation}
D(S) \;=\; 1 - \bar{s}(S), \qquad \bar{s}(S) \;=\; \frac{2}{|S|(|S|-1)} \sum_{i<j} s(e_i,e_j).
\end{equation}

$D(S)\in[0,1]$ is larger when exemplars are linguistically diverse (lower average similarity).

\paragraph{Combined objective (\emph{R-score}).}
We combine label- and text-diversity via a convex mixture
\begin{equation}
\label{eq:Rscore}
R(S) \;=\; \alpha \, G(S) \;+\; (1-\alpha)\, D(S), \qquad \alpha\in[0,1].
\end{equation}
where $\alpha$ balances category coverage ($G$) and linguistic variety ($D$). $\alpha{=}1$ emphasizes intent coverage only; $\alpha{=}0$ emphasizes linguistic variety only.

\paragraph{Constrained subset selection.}
Given pool $S_L$, we select a size-$K$ subset $T_K$ by
\begin{align}
\label{eq:opt}
T_K = \arg\max_{S\subseteq S_L,\; |S|=K} \; R(S) 
\quad \text{s.t.} \quad 
\begin{aligned}
& s(z_n,e_i) \ge \tau, && \forall i \in S, \\
& n_k(S) \le U, && \forall k.
\end{aligned}
\end{align}
where $\tau$ is a minimum relevance threshold to $z_n$, and $U$ limits per-label multiplicity to avoid collapse.
$K$ is the number of exemplars to include; $\tau$ preserves relevance; $U$ prevents one label dominating.

\paragraph{Fast greedy selection with closed-form increments.}
We employ a greedy algorithm with marginal gain:
\begin{equation}
\Delta R(i \mid S) \;=\; R(S\cup\{i\}) - R(S),
\end{equation}
which can be computed in $O(|S|)$ per candidate using closed-form updates.
Let $m=|S|$, $c=\sum_k p_k^2$, and $A=\sum_{j\in S} s(e_i,e_j)$. Then
\begin{align}
\label{eq:gini_update}
G(S\cup\{i\}) &= 1 - \sum_k \Big(\frac{n_k + \mathbb{I}[y_i=k]}{m+1}\Big)^2 \nonumber \\
&= 1 - \frac{1}{(m+1)^2}\!\left(\sum_k n_k^2 + 2 n_{y_i} + 1\right), \\
\label{eq:text_update}
\bar{s}(S\cup\{i\}) &= \frac{\binom{m}{2}\,\bar{s}(S) + A}{\binom{m+1}{2}}
\;\;\Rightarrow\;\;
D(S\cup\{i\}) = 1 - \bar{s}(S\cup\{i\}).
\end{align}
Thus $\Delta G,\Delta D$ follow from \eqref{eq:gini_update}--\eqref{eq:text_update} and $\Delta R$ from \eqref{eq:Rscore}. We optionally add a relevance prior to break ties:
\begin{equation}
\widetilde{\Delta R}(i\mid S) \;=\; \Delta R(i\mid S) \;+\; \mu \, s(z_n,e_i), \quad \mu\ge 0.
\end{equation}
where $n_{y_i}$ is the count of label $y_i$ in $S$; $A$ accumulates pairwise similarities to $i$.

MMR\citep{xia2015learning} selects $i$ by $\lambda s(z_n,e_i) - (1-\lambda)\max_{j\in S}s(e_i,e_j)$.  FPS\citep{moenning2003fast} picks the farthest item to $S$.
Our $R$ jointly optimizes label coverage and average set-wise dissimilarity with explicit constraints (Eq.\ref{eq:opt}), which we find more faithful to multi-intent dialogue.

\subsection{Instruction Prompt Synthesis}
LLMs benefit from structured, compact prompts that (i) summarize context, (ii) present diverse, relevant exemplars, and (iii) request a single intent decision with explanation.
We compose a prompt $\Pi$ with slots:
\begin{equation}
\label{eq:pi}
\begin{aligned}
\Pi = \big[& \texttt{System Instruction}; \\
& \texttt{Context Summary}(C_{1:n-1}); \\
& \texttt{Current Utterance: }q_n; \\
& \texttt{Exemplars: }\{(x_i,y_i)\}_{i\in T_K}; \\
& \texttt{Answer format} \big].
\end{aligned}
\end{equation}
A lightweight summarizer distills $C_{1:n-1}$ into $\sigma(C_{1:n-1})$ (token budgeted). Exemplars are rendered as pairwise (query, intent) lines with minimal boilerplate to fit the context window.

\subsection{LLM-Based Intent Decoding and Calibration}
Direct generation can be verbose, we prefer calibrated scores over candidate intents and a deterministic decision.

Let $\Phi$ be a fixed LLM. We score an intent $y$ with a discriminative likelihood:
\begin{equation}
\label{eq:score}
\begin{aligned}
S_\Phi(y \mid \Pi) = \log &P_\Phi(\texttt{<yes>} \mid \Pi, \texttt{``Is intent = }y\texttt{?''}) \\
- \log &P_\Phi(\texttt{<no>} \mid \Pi, \texttt{``Is intent = }y\texttt{?''}),
\end{aligned}
\end{equation}
and predict $\hat{y}=\arg\max_{y\in \mathcal{Y}_\Pi} S_\Phi(y\mid \Pi)$, where $\mathcal{Y}_\Pi$ is the set of unique labels exposed via $T_K$ (plus a small shortlist from retrieval to mitigate coverage loss). We apply temperature scaling $S_\Phi/\tau_c$ to calibrate confidence. 
$S_\Phi$ is a log-odds style verifier, $\tau_c>0$ is a calibration temperature.

\subsection{Learning, Hyperparameters, and Efficiency Budget}
\label{sec:learning_hparam_efficiency}
LDRA is trained and tuned under a deployment-aware regime: we learn discriminative encoders and select diversification knobs while explicitly respecting an end-to-end latency budget. Concretely, let $\mathcal{D}=\{(x,y)\}$ denote a small set of labeled single-turn pairs and $\mathcal{M}$ the retrieval memory (with precomputed embeddings $e_i$). At inference on a dialogue $(C_{1:n-1},q_n)$, the wall-clock latency decomposes into four terms
\begin{equation}
\label{eq:latency_decomp}
\underbrace{t_{\text{ANN}}}_{\substack{\text{vector/}\\\text{lexical retrieval}}} \;+\;
\underbrace{t_{\text{DIV}}}_{\text{diversification}} \;+\;
\underbrace{t_{\text{PROMPT}}}_{\substack{\text{context}\\\text{summary}\,+\,\\\text{compose}}} \;+\;
\underbrace{t_{\text{LLM}}}_{\substack{\text{decode/}\\\text{verify}}}
\;\;\le\;\; \mathsf{B},
\end{equation}
where $\mathsf{B}$ is a given budget (e.g., voice-bot real-time). We therefore optimize a constrained objective:
\begin{equation}
\label{eq:constrained_obj}
\max_{\Theta} \;\; \mathrm{Acc}_{\text{dev}}(\Theta)
\quad\text{s.t.}\quad
\mathbb{E}[t_{\text{ANN}}{+}t_{\text{DIV}}{+}t_{\text{PROMPT}}{+}t_{\text{LLM}}]\le \mathsf{B},
\end{equation}
with $\Theta=\{f_\theta,\alpha,\tau,U,L,K,\lambda_{\text{vec}},\mu,\tau_c,\ldots\}$ collecting trainable and tunable components.
$\alpha$ is diversity weight in $R(S)$, $\tau$ relevance threshold, $U$ per-label cap, $L$ pool size, $K$ exemplar count, $\lambda_{\text{vec}}$ hybrid-retrieval weight in Eq.\eqref{eq:hybrid}, $\mu$ relevance tie-breaker, $\tau_c$ calibration temperature).

\paragraph{Encoder learning (metric fine-tuning).}
We fine-tune the encoder $f_\theta$ on $\mathcal{D}$ to produce embeddings that (i) cluster same-intent texts and (ii) separate different-intent texts, which improves both retrieval precision and the \emph{text-diversity} $D(\cdot)$ signal:
\begin{equation}
\label{eq:metric_learning}
\begin{aligned}
\mathcal{L}_{\text{metric}} = \sum_{(u,v)}
\big[& \mathbb{I}[y_u{=}y_v]\,(1-s(e_u,e_v))_+ \\
&+ \mathbb{I}[y_u{\neq}y_v]\,(s(e_u,e_v){-}m)_+ \big].
\end{aligned}
\end{equation}
where $e_u{=}f_\theta(u)$, $s(\cdot,\cdot)$ is cosine similarity, $(\cdot)_+=\max(\cdot,0)$, and $m\!\in\!(0,1)$ is a margin. In practice, we strengthen Eq.\eqref{eq:metric_learning} with hard negative mining: for each anchor $u$, we retrieve near neighbors by ANN and re-label those with $y{\neq}y_u$ as hard negatives, this accelerates large-margin separation.

\paragraph{Diversity and retrieval knobs (black-box tuning).}
The selection in Eq.\eqref{eq:opt} is discrete, we therefore treat $(\alpha,\tau,U,L,K,\lambda_{\text{vec}},\mu)$ as black-box hyperparameters and search them on a dev split with the latency constraint in Eq.\eqref{eq:constrained_obj}. Two equivalent views are useful:
\begin{align}
\textbf{(Constrained)}\quad&
\max_{\Theta}\;\mathrm{Acc}_{\text{dev}}(\Theta)
\;\text{s.t.}\;
\mathbb{E}[t_{\text{tot}}(\Theta)]\le \mathsf{B}, \\
\textbf{(Scalarized)}\quad&
\max_{\Theta}\; \mathrm{Acc}_{\text{dev}}(\Theta)
-\lambda_{\ell}\,\max\!\Big\{0,\frac{\mathbb{E}[t_{\text{tot}}(\Theta)]}{\mathsf{B}}-1\Big\},
\end{align}
where $t_{\text{tot}}$ refers to the left-hand side of \eqref{eq:latency_decomp} and $\lambda_{\ell}\!>\!0$ is a penalty weight. We adopt small bounded grids (e.g., $\alpha\!\in\![0.2,0.8]$, $\tau\!\in\![0.2,0.6]$, $U\!\in\!\{1,2\}$, $L\!\in\!\{64,128,256\}$, $K\!\in\!\{4,6,8\}$, $\lambda_{\text{vec}}\!\in\![0.4,0.8]$, $\mu\!\in\![0,0.2]$) and Bayesian search for fine refinement. 

\paragraph{Calibration and distillation.}
Given a structured prompt $\Pi$, we compute a verifier log-odds $S_\Phi(y\mid\Pi)$ (Eq.~\eqref{eq:score}) and produce calibrated probabilities
\(
p_\Phi(y\mid\Pi)=\sigma(S_\Phi(y\mid\Pi)/\tau_c)
\)
with temperature $\tau_c{>}0$ (\emph{symbol:} $\sigma$ is the logistic). For low-latency deployment, we distill to a compact classifier $\psi$:
\begin{equation}
\label{eq:distill}
\mathcal{L}_{\text{distill}}
\;=\;
-\sum_{(C,q_n)} \sum_{y\in\mathcal{Y}_\Pi}
\underbrace{\mathrm{softmax}\!\big(S_\Phi(y\mid\Pi)/\tau_c\big)}_{\text{teacher target}}
\log \underbrace{\mathrm{softmax}\!\big(z_\psi(C,q_n)_y\big)}_{\text{student logits}},
\end{equation}
keeping LDRA retrieval \& diversification \emph{unchanged} so the student inherits both accuracy and diversity benefits. (We also apply a margin-based \emph{pairwise} distillation between the top-1 and non-top intents to preserve teacher ranking.)

\paragraph{Token/latency accounting and complexity.}
We express each term in Eq.\eqref{eq:latency_decomp} with dominant costs and symbols defined locally to avoid cross-reference:
\begin{align}
\label{eq:t_ann}
t_{\text{ANN}} &\approx c_{\text{ann}}\log N + c_{\text{bm25}}\,\#\text{terms}(q_n), \\
&\quad \text{(ANN/HNSW or IVF)} \nonumber \\
\label{eq:t_div}
t_{\text{DIV}} &\approx c_{\text{sim}}\,(LK) + c_{\Delta}\,K, \\
&\quad \text{(greedy $K$ steps, each scans $\le L$ candidates)} \nonumber \\
\label{eq:t_prompt}
t_{\text{PROMPT}} &\approx c_{\text{sum}}\,|C| + c_{\text{fmt}}\,K, \\
&\quad \text{(lightweight history summarizer + exemplar rendering)} \nonumber \\
\label{eq:t_llm}
t_{\text{LLM}} &\approx \frac{|\Pi| + T_{\text{gen}}}{r_{\text{tok}}}, \\
&\quad \text{(LLM throughput $r_{\text{tok}}$ tokens/s)} \nonumber
\end{align}
where $N$ memory size; $c_{\star}$ hardware/index constants, $|C|$ number of turns, $|\Pi|$ prompt tokens, $T_{\text{gen}}$ generated tokens until decision, $r_{\text{tok}}$ decoding rate. 
On the algorithmic side, with embeddings precomputed, similarity retrieval is $O(\log N)$ per query (index-dependent), yielding $O(LK)$ new similarities overall (we maintain running sums $A_i{=}\sum_{j\in S}s(e_i,e_j)$ to avoid recomputing all pairs).
Prompt construction is linear in the context size and $K$.
Thus, the overall per-query complexity is
\begin{equation}
\label{eq:overall_complexity}
\mathcal{T}_{\text{LDRA}} \;=\; O(\log N) \;+\; O(LK) \;+\; O(|C|{+}K) \;+\; O(|\Pi|{+}T_{\text{gen}}),
\end{equation}
which maps directly to Eq.\eqref{eq:latency_decomp} via the constants in Eq.\eqref{eq:t_ann}--\eqref{eq:t_llm}.

\paragraph{Adaptive budget control .}
We expose two anytime levers to respect $\mathsf{B}$ at runtime without re-tuning:
(i) \textbf{Pool shrinking}: reduce $L$ (and proportionally $K$) when instantaneous load increases; 
(ii) \textbf{Prompt compression}: shorten $\sigma(C_{1:n-1})$ and cap $|\Pi|$; both lower $t_{\text{PROMPT}}$ and $t_{\text{LLM}}$. 
A simple controller enforces $t_{\text{tot}}\!\le\!\mathsf{B}$ by greedily decreasing $(L,K)$ while preserving per-label cap $U$ to retain label coverage.

LDRA couples a context encoder with a principled diversification objective $R(S)=\alpha G(S)+(1-\alpha)D(S)$ to select supervision exemplars that jointly maximize intent coverage and linguistic variety, and then leverages an LLM verifier for calibrated intent decoding.

\section{Experiment}
\subsection{Experimental Setup}
\subsubsection{Datasets}
\label{sec:datasets}
We evaluate LDRA on widely–adopted, multi–domain, task–oriented dialogue benchmarks that match our problem setting (multi–turn intent/state understanding) and are commonly reported by recent SOTA and baseline systems.

\paragraph{SGD (Schema-Guided Dialogue).}
A large-scale, multi-domain corpus (about 16k dialogues) spanning 16 domains and 26 services (e.g., restaurant, flight, hotel, media, payment)\citep{rastogi2020towards}. We use the official train/dev/test splits and the released schemas. For our retrieval memory $\mathcal{M}$ we index user turns paired with their gold domain–slot annotations, labels are normalized into \texttt{not\_mentioned} / canonical values.

\paragraph{MultiWOZ.}
A fully annotated, multi-domain corpus (about 10k dialogues) across 8 domains (restaurant, hotel, attraction, taxi, train, hospital, police, bus)\citep{budzianowski2018multiwoz}. We report on \emph{MultiWOZ~2.1} and \emph{MultiWOZ~2.4}. We follow the community preprocessing for each version and keep the official splits. For LDRA, each user turn contributes an exemplar \((x_i, y_i)\) where $x_i$ is the turn text and $y_i$ is the normalized state (active domain(s) and slot–value map) at that turn.

On both datasets we evaluate multi–turn state/intent understanding in the standard Dialogue State Tracking (DST) formulation. Concretely, LDRA receives the dialogue history $C_{1:n-1}$ and current user utterance $q_n$, retrieves diverse exemplars from $\mathcal{M}$, and predicts the current state (domain(s) and slot values). This directly supports joint comparisons with DST baselines.

\newcommand{\best}[1]{\cellcolor{gray!30}\textbf{#1}}
\newcommand{\second}[1]{\cellcolor{gray!18}\textbf{#1}}
\newcommand{\third}[1]{\cellcolor{gray!10}\textbf{#1}}

\begin{table}[t]
\centering
\small
\setlength{\tabcolsep}{6pt}
\renewcommand{\arraystretch}{1.15}
\caption{\textbf{Overall JGA (\%)} on MultiWOZ~2.0/2.4 and SGD. Best, second-best, and third-best cells are shaded in 
\colorbox{gray!30}{\strut \textbf{dark}}/\colorbox{gray!18}{\strut \textbf{medium}}/\colorbox{gray!10}{\strut \textbf{light}} gray, respectively.}
\label{tab:overall_jga}
\begin{tabular}{lccc}
\toprule
\textbf{Model} & \textbf{MultiWOZ 2.0} & \textbf{MultiWOZ 2.4} & \textbf{SGD} \\
\midrule
TRADE & 48.62 & 55.10 & -- \\
T5DST & 53.42 & -- & -- \\
DiSTRICT & 57.02 & -- & -- \\
D3ST & 58.60 & 75.90 & \third{86.40} \\
LDST–llama-7b & 60.65 & 79.94 & 84.47 \\
ChatGPT-4o & \third{64.30} & \third{83.20} & 84.80 \\
LLaMA-7b (prompted) & 55.37 & 75.13 & 75.32 \\
IC-DST Codex & -- & -- & 62.40 \\
RefPyDST & -- & -- & 65.20 \\
\midrule
\textbf{LDRA–LoRA (ours)} & \second{67.34} & \second{85.57} & \second{89.04} \\
\textbf{LDRA–GRPO (ours)} & \best{70.34} & \best{89.35} & \best{92.01} \\
\bottomrule
\end{tabular}
\end{table}
\vspace{-2mm}

\subsubsection{Evaluation Metrics}
\label{sec:metrics}
To align with prior work, we adopt:
\begin{itemize}
    \item \emph{Joint Goal Accuracy (JGA)}: fraction of turns for which the \emph{entire} predicted dialogue state exactly matches the gold state. This is the primary metric we report.
    \item \emph{Average Goal Accuracy (AGA)}: per–turn average of slot-wise correctness over active slots. We compute AGA for completeness but, consistent with prior observations, it is often saturated; hence we mainly analyze JGA.
\end{itemize}
Unless otherwise noted, we report mean~$\pm$~std over 3 runs (API models are sampled 3 times with identical prompts, trainable models are run with 3 random seeds).

\subsubsection{Baselines and Compared Systems}
\label{sec:baselines}
We compare LDRA with strong DST systems and recent LLM-based approaches. The list covers both classical encoders and instruction-following LLMs widely used in current SOTA comparisons:

(i) \textit{Encoder/seq2seq DST baselines.}
\textbf{TRADE}\citep{wu2019transferable}, \textbf{T5DST} \citep{lin2021leveraging}, \textbf{DiSTRICT}\citep{venkateswaran2023district}, \textbf{D3ST} \citep{zhao2022description}, \textbf{IC-DST}\citep{hu2022context}, \textbf{RefPyDST}\citep{king2023diverse}, \textbf{SM2-11B}\citep{chen2023stabilized}, \textbf{LDST-llama-7b}\citep{feng2023towards}, \textbf{LLaMA-7B} (prompted).

(ii) \textit{LLM baselines (prompted / API)}
\textbf{ChatGPT-4o}, \textbf{Gemini-2.5}, \textbf{Doubao-1.5}, \textbf{DeepSeek-v3.2} (API). For fair comparison, all APIs use the same instruction template and constrained answer format; temperature is set to $0.2$ and top-$p$ to $0.95$ unless otherwise stated.

(iii) \textit{Open-weight chat models.}
\textbf{Qwen} (our primary family), and \textbf{Kimi}. We also report numbers from related papers when official checkpoints or evaluation scripts are unavailable, citing the source.

Please refer to \S\ref{sec:impl} for the detailed implementations.

\begin{table}[ht]
\centering
\small
\setlength{\tabcolsep}{4pt}
\renewcommand{\arraystretch}{1.1}
\caption{\textbf{Detailed Performance Analyses on MultiWOZ~2.4}}
\label{tab:breakdown}
\begin{tabular}{@{}lccccc@{}}
\toprule
\multicolumn{6}{c}{\textbf{(A) Domain-wise Joint Goal Accuracy on MultiWOZ~2.4}} \\
\midrule
\textbf{Model} & \textbf{Attract.} & \textbf{Hotel} & \textbf{Rest.} & \textbf{Taxi} & \textbf{Train} \\
\midrule
TRADE & 21.56 & 14.57 & 14.01 & 61.02 & 23.56 \\
T5DST & 33.97 & 23.81 & 23.46 & 65.02 & 35.38 \\
DiSTRICT & 32.98 & 22.28 & 24.99 & 66.86 & 47.24 \\
SUMBT & 22.97 & 20.33 & 16.95 & 61.03 & 23.01 \\
SimpleTOD & 29.08 & 18.29 & 16.18 & 62.00 & 28.96 \\
D3ST & 57.29 & 23.07 & 38.79 & 79.83 & 40.80 \\
IC-DST Codex & \third{62.10} & \third{53.20} & \third{54.90} & \third{71.90} & \third{51.40} \\
RefPyDST & \best{74.50} & \second{56.60} & \second{68.20} & 68.50 & \best{76.10} \\
\midrule
\textbf{LDRA (ours)} & \second{72.99} & \best{68.38} & \best{73.28} & \best{84.08} & \second{75.57} \\
\bottomrule
\end{tabular}

\vspace{2em}

\begin{tabular}{@{}lccc@{}}
\toprule
\multicolumn{4}{c}{\textbf{(B) Few-shot Joint Goal Accuracy on MultiWOZ~2.4}} \\
\midrule
\textbf{Model} & \textbf{1\%} & \textbf{5\%} & \textbf{10\%} \\
\midrule
DS2-BART & 30.6 & 42.5 & 41.7 \\
DS2-T5 & 36.8 & 49.9 & 51.1 \\
IC-DST GPT-Neo & 17.4 & 29.6 & 34.4 \\
IC-DST CodeGen & 21.9 & 33.2 & 37.5 \\
IC-DST Codex & 48.4 & 55.4 & 56.9 \\
SM2-11B & 40.0 & 51.1 & 52.0 \\
LDST & 46.8 & 56.5 & \third{62.5} \\
RefPyDST & \third{49.4} & \third{57.9} & 60.57 \\
\midrule
\textbf{LDRA–LoRA (ours)} & \best{55.34} & \second{74.58} & \second{78.78} \\
\textbf{LDRA–GRPO (ours)} & \second{50.78} & \best{77.03} & \best{79.92} \\
\bottomrule
\end{tabular}

\vspace{2em}

\begin{tabular}{@{}lcc@{}}
\toprule
\multicolumn{3}{c}{\textbf{(C) Cross-dataset Transfer (train$\rightarrow$test)}} \\
\midrule
\textbf{Model} & \textbf{SGD$\rightarrow$MWOZ2.4} & \textbf{MWOZ2.4$\rightarrow$SGD} \\
\midrule
D3ST & 28.9 & 23.1 \\
LDST & 31.6 & 25.9 \\
\textbf{LDRA–LoRA (ours)} & \second{79.39} & \second{87.02} \\
\textbf{LDRA–GRPO (ours)} & \best{86.89} & \best{89.37} \\
\bottomrule
\end{tabular}
\end{table}

\subsection{Main Results}
\label{sec:main_results}
We evaluate four settings: (i) zero-shot prompting with RAG, (ii) few-shot SFT with RAG (\texttt{LDRA–LoRA}), (iii) few-shot RL with RAG (\texttt{LDRA–GRPO}), and (iv) full-data (when applicable). We report Joint Goal Accuracy (JGA) as the primary metric and provide domain-wise and data-efficiency analyses.

\paragraph{Overall.}
As shown in \autoref{tab:overall_jga}, LDRA establishes new SOTA on all three benchmarks. 
On MultiWOZ~2.0, \texttt{LDRA–GRPO} reaches \best{70.34} JGA, surpassing the strongest baseline (ChatGPT-4o, \third{64.30}) by \(+6.0\) points; \texttt{LDRA–LoRA} already outperforms all baselines at \second{67.34}. 
On MultiWOZ~2.4, \texttt{LDRA–GRPO} scores \best{89.35} ( \(+6.15\) over ChatGPT-4o), with \texttt{LDRA–LoRA} at \second{85.57}. 
On SGD, \texttt{LDRA–GRPO} achieves \best{92.01} and \texttt{LDRA–LoRA} \second{89.04}, improving over the best encoder baseline D3ST (\third{86.40}) by \(+5.61\) / \(+2.64\) points. 
These gains persist across datasets and training regimes, indicating that diversity-aware retrieval benefits both SFT and RL finetuning.

\paragraph{Domain breakdown.}
From \autoref{tab:breakdown}(A), LDRA delivers the best average JGA on MultiWOZ~2.4 and is top on 3/5 domains: \emph{Hotel} (+11.78 over the best baseline), \emph{Restaurant} (+5.08), and \emph{Taxi} (+4.25). It is the runner-up on \emph{Attraction} (–1.51 vs.\ RefPyDST) and \emph{Train} (–0.53 vs.\ RefPyDST). The gains are consistent with LDRA’s intent-coverage and linguistic-variety objectives, which particularly help domains with heterogeneous slot realizations (e.g., \emph{Hotel}, \emph{Restaurant}).

\paragraph{Data efficiency.}
As shown in \autoref{tab:breakdown}(B), LDRA is highly sample-efficient. With only 1\% training data, \texttt{LDRA–LoRA} attains 55.34 JGA (best) and \texttt{LDRA–GRPO} 50.78, both outperforming strong few-shot baselines (e.g., RefPyDST at \third{49.4}, Codex at 48.4). At \textbf{5\%} and \textbf{10\%} data, \texttt{LDRA–GRPO} remains best with 77.03 and 79.92, respectively, indicating that retrieval diversity continues to matter even as data increases.

\paragraph{Cross-dataset generalization.}
\autoref{tab:breakdown}(C) demonstrates robust transfer. Training on SGD and testing on MultiWOZ~2.4, \texttt{LDRA–GRPO} achieves 86.89 JGA, far exceeding encoder baselines (e.g., LDST at 31.6). The reverse transfer (MultiWOZ~2.4$\rightarrow$SGD) is similarly strong. These results suggest LDRA’s diverse exemplars reduce distribution shift by covering alternative surface forms and intent realizations.

Across benchmarks, domains, and data regimes, LDRA consistently improves JGA—often by \(+4\sim6\) absolute over the strongest published baselines—while maintaining strong transfer. The performance trend from zero-/few-shot to GRPO further indicates that diversity-aware retrieval composes well with both SFT and RL training.

\begin{figure*}[ht]
  \centering
  \includegraphics[width=\linewidth]{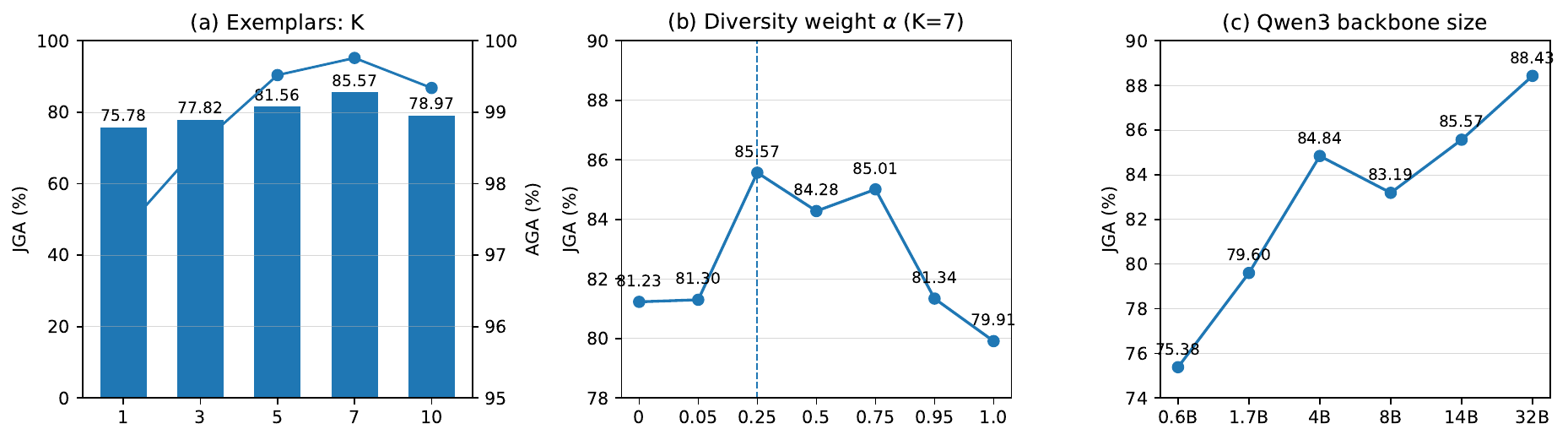}
  \caption{\textbf{Parameter sensitivity on MultiWOZ~2.4.} 
  \textbf{(a)} Effect of exemplar count $K$: JGA (bars) and AGA (line); JGA peaks at $K{=}7$, after which prompt redundancy/length hurts.
  \textbf{(b)} Effect of $\alpha$ with $K{=}7$: a balanced mixture is best; $\alpha{=}0.25$ attains the highest JGA.
  \textbf{(c)} Qwen3 scaling: larger backbones generally help; LDRA benefits across scales (0.6B$\rightarrow$32B).}
  \label{fig:sensitivity_overview}
\end{figure*}

\subsection{Ablation and Analysis}
\label{sec:ablation}
All ablations are conducted on MultiWOZ~2.4 with ground-truth domains.
The retrieval memory $\mathcal{M}$ is constructed only from the training split.
\subsubsection{Overall Structure}

\begin{figure}[ht]
    \centering
    \includegraphics[width=0.7\linewidth]{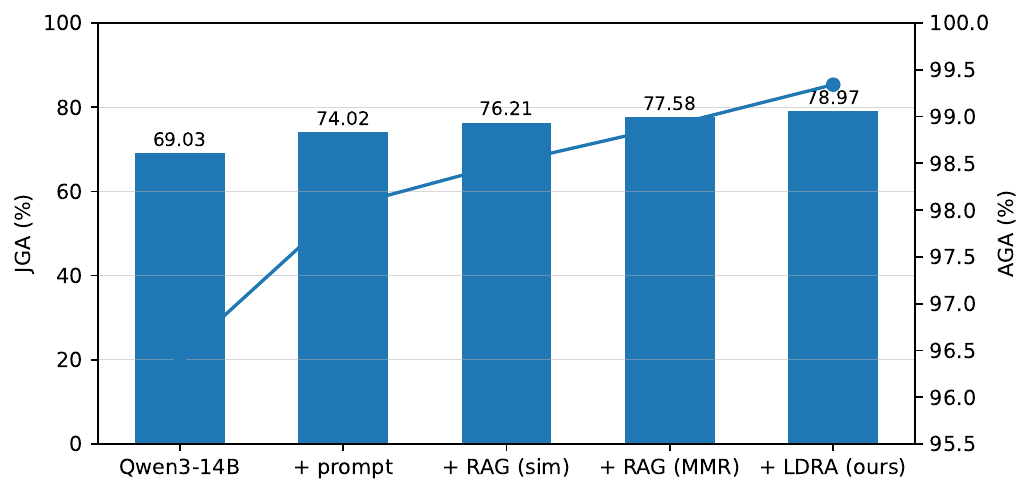}
    \caption{\textbf{Overall structure ablation on MultiWOZ~2.4.} 
    Bars show JGA; the curve shows AGA. 
    Starting from the base model (Qwen3-14B), adding a task instruction yields a strong gain; 
    similarity-only RAG and MMR bring incremental improvements; 
    LDRA (ours) further increases JGA by promoting label coverage and linguistic variety in retrieved exemplars.}
    \label{fig:overall_structure_mwoz24}
\end{figure}

As visualized in \autoref{fig:overall_structure_mwoz24}, 
a task-specific instruction already lifts JGA substantially (\(+4.99\) points over the base). 
Adding similarity-only RAG helps further, and MMR adds another modest gain by removing redundancy. 
Our full LDRA brings the largest single improvement over the prompted baseline (\(+4.95\) points), 
while AGA is near-saturated across strong variants. These results confirm that JGA is the discriminative metric in this regime.

\subsubsection{Parameter Sensitivity}
\label{sec:sensitivity}

We probe three knobs: (a) the number of retrieved exemplars $K$ used in prompting, (b) the diversity weight $\alpha$ in $R(S)=\alpha G(S)+(1-\alpha)D(S)$, and (c) the backbone size of Qwen3. All runs use ground-truth domains and a retrieval memory built only from the training split.

Fig.~\ref{fig:sensitivity_overview} shows (i) increasing $K$ improves JGA up to $K{=}7$ (85.57) before degrading at $K{=}10$ (78.97), (ii) $\alpha{=}0.25$ maximizes JGA (85.57), while pure $G$ or pure $D$ underperform, and (iii) JGA improves with model scale, reaching 88.43 on Qwen3-32B.

\subsubsection{Necessity of $R = \alpha G + (1 - \alpha) D$}
\label{sec:ablation_alpha}
Do gains come from \emph{label coverage} ($G$) or textual diversity ($D{=}1{-}\mathrm{Sim}$)?  
We sweep $\alpha\!\in\!\{0,\,0.25,\,0.5,\,0.75,\,1\}$ under fixed $K, L, \tau, U$; retrieval memory uses the training split only (no dev/test leakage).

We find that:

(1) The JGA heatmap shows a \emph{bimodal sweet spot} in $\alpha$: mixtures ($0.25{\sim}0.75$) outperform either component alone across $K$, indicating that \emph{both} label coverage and linguistic variety matter.  

(2) At K=7, radar profiles reveal that pushing only $G$ or only $D$ skews the set geometry and does not yield the best JGA, balanced trade-offs maximize $R$ and align with higher accuracy.

(3) Across $K$, increasing exemplars beyond the sweet spot (e.g., $K{=}10$) reduces JGA even when $G$ grows, due to redundancy/longer prompts (higher similarity), emphasizing the need to jointly control $G$ and $D$.

\subsubsection{Diversity Objective vs.\ Classic Re-ranking }
\label{sec:diversity_vs_classics}
We fix the \emph{token budget} and the number of exemplars \(K\) \emph{identically} across all methods, and compare:
(i) Similarity Top-\(K\),
(ii) MMR re-ranking,
(iii) FPS (farthest-point) selection,
(iv) LDRA with label-only \(G\),
(v) LDRA with text-only \(D\),
(vi) LDRA (full) with \(R(S)=\alpha G(S)+(1-\alpha)D(S)\).

\begin{figure}[ht]
  \centering
  \includegraphics[width=0.6\linewidth]{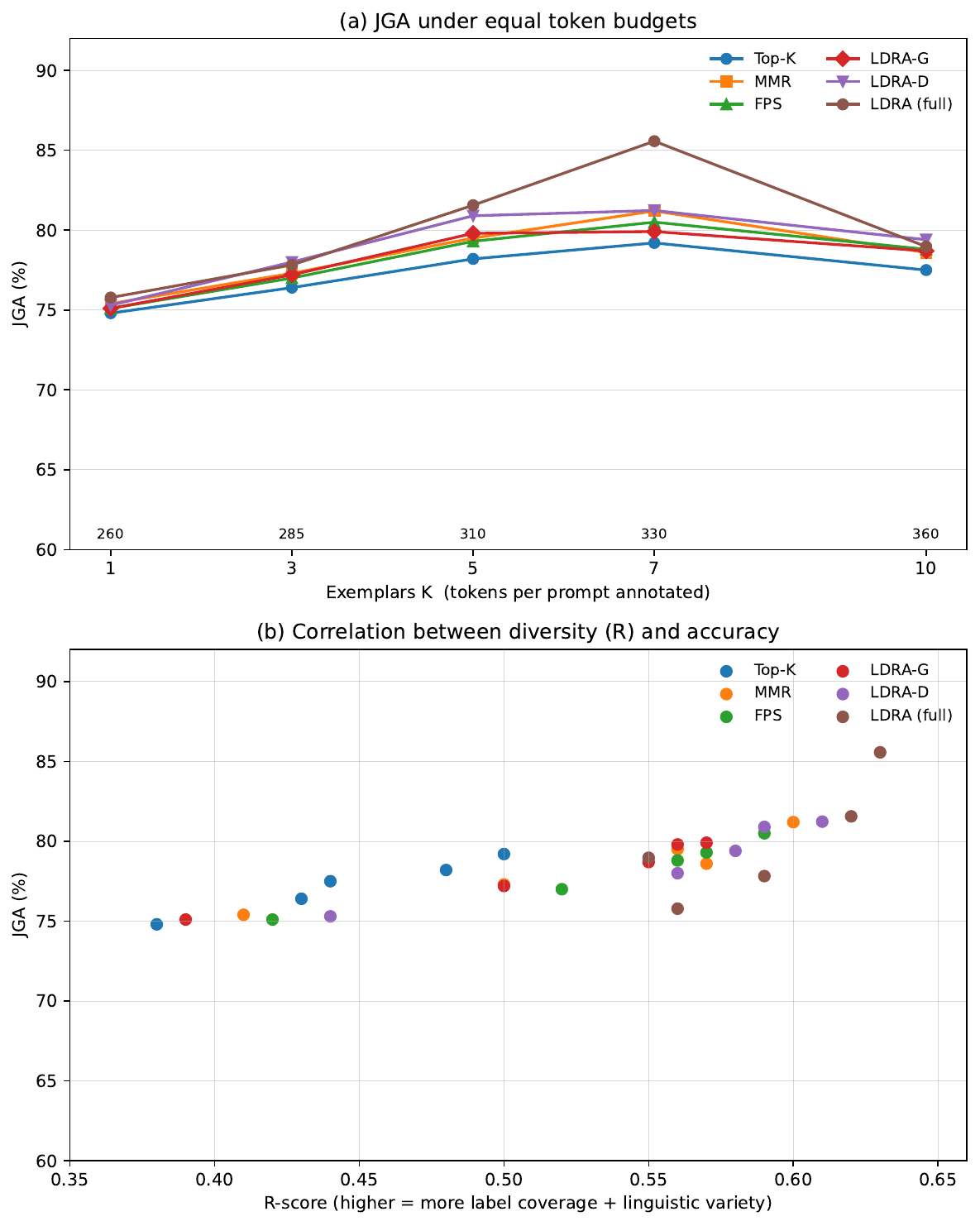}
  \caption{\textbf{Diversity vs.\ classic re-ranking under equal token budgets.}
  \textbf{(Up)} JGA vs.\ \(K\in\{1,3,5,7,10\}\): LDRA (full) dominates and peaks at \(K{=}7\); MMR/FPS improve over similarity Top-\(K\).
  \textbf{(Down)} \(R\)-score vs.\ JGA across methods and \(K\) shows a strong positive trend, indicating that sets with higher label coverage and linguistic variety translate into higher accuracy.}
  \label{fig:diversity_vs_classics}
\end{figure}

\begin{figure}[ht]
  \centering
  \includegraphics[width=0.4\linewidth]{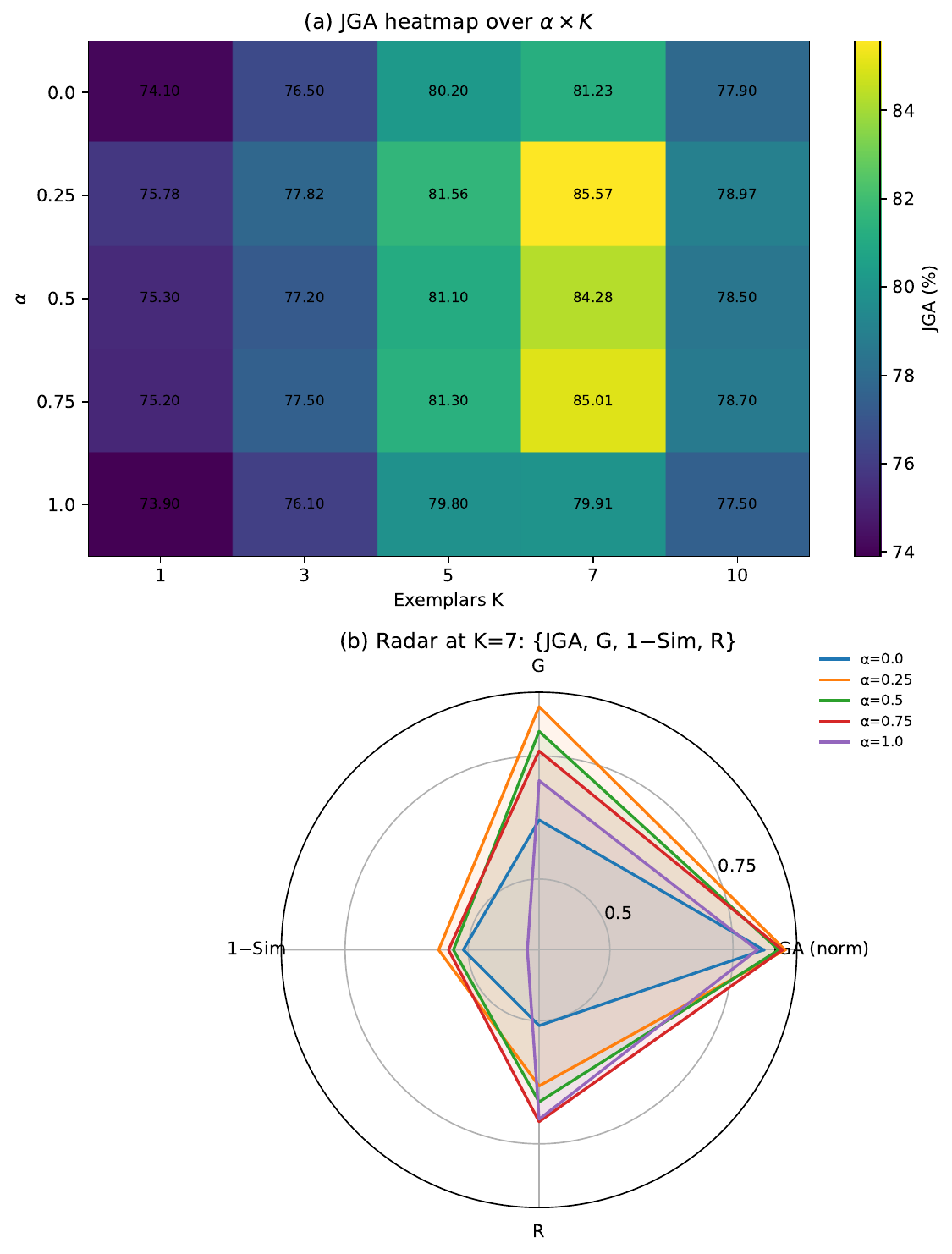}
  \caption{\textbf{Effect of $\alpha$ (MultiWOZ~2.4).} 
  \textbf{(Up)} Heatmap of JGA across $\alpha\times K$ (token budget and $K$ matched across methods).
  JGA peaks for a balanced mixture (e.g., $\alpha\!\approx\!0.25$–$0.75$) and saturates or declines at extremes ($\alpha{=}0$/$1$). 
  \textbf{(Down)} Radar profiles at $K{=}7$ contrasting $\{\text{JGA},\,G,\,1{-}\mathrm{Sim},\,R\}$ over $\alpha$ (metrics normalized to $[0,1]$ for display; $R$ is computed as $R=\alpha G+(1-\alpha)(1-\mathrm{Sim})$.}
  \label{fig:alpha_ablation_overview}
\end{figure}

\begin{figure}[ht]
  \centering
  \includegraphics[width=0.4\linewidth]{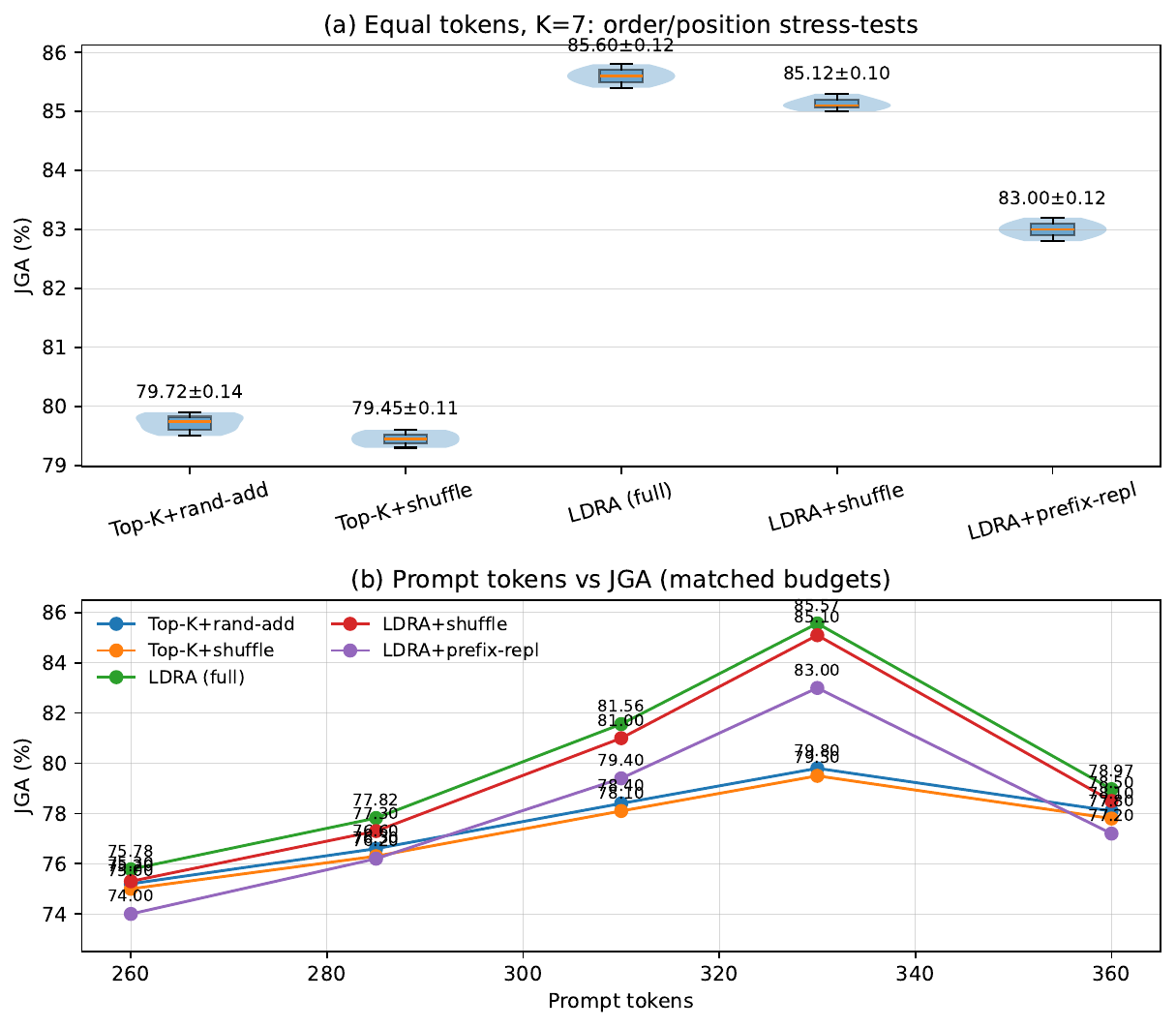}
  \caption{\textbf{Token budget \& position bias controls.}
  \textbf{(Top)} Distributions of JGA under \emph{equal tokens, $K{=}7$}: LDRA remains substantially above matched-token baselines; shuffling has minor effect, while replacing the \emph{prefix} harms performance (LDRA$>$LDRA+shuffle$\gg$LDRA+prefix-repl \& Top-$K$ variants). 
  \textbf{(Bottom)} \emph{Prompt tokens vs JGA} with token budgets matched at $\{260,285,310,330,360\}$: LDRA dominates at each budget; simply adding random exemplars (Top-$K$+rand-add) increases tokens without closing the JGA gap.}
  \label{fig:token_position_fairness}
\end{figure}

From Fig.~\ref{fig:diversity_vs_classics}, LDRA (full) consistently outperforms classic re-ranking under the \emph{same} \(K\) and prompt length, peaking at \(K{=}7\) (85.57 JGA).  
MMR/FPS yield smaller, stable gains over similarity Top-\(K\), but fall short of LDRA.  
The right panel reveals a clear correlation between the combined diversity score \(R\) and JGA across methods/\(K\): better label coverage together with lower redundancy yields higher accuracy.

\subsubsection{Fairness Control: Token Budget \& Position Bias}
\label{sec:fairness_token_position}
Are LDRA’s gains merely due to more context or better exemplar positions?  
We match the prompt token budget across methods and fix $K,L,\tau,U$. As a fairness control, we build a baseline by adding random exemplars to the Top-$K$ retriever to reach the same token budget as LDRA (``Top-$K$+rand-add''). We then stress-test position bias via (i) shuffle (randomly permute exemplar order per run), and (ii) prefix-replace (replace the first half exemplars with randomly selected ones of matched label distribution).

It can be found from Fig.\ref{fig:token_position_fairness} that:
(\emph{i}) With \emph{equal tokens} and fixed $K{=}7$, LDRA’s JGA distribution is well-separated from the Top-$K$ controls; shuffling order has negligible impact, confirming LDRA is not exploiting a fixed position prior. 
(\emph{ii}) \emph{Prefix replacement} degrades LDRA notably, indicating early exemplars do matter \emph{if they are informative and diverse}.  
(\emph{iii}) As prompt tokens increase, LDRA’s advantage persists—``more tokens'' via random additions does not match \emph{diversity-aware} retrieval.

\vspace{3mm}

\section{Conclusion}
In this paper, we addressed whether retrieval diversity—balancing intent coverage and linguistic variety—can reliably improve LLM-based multi-turn intent understanding under fixed token budgets, and introduced a diversity-aware selection framework with budget- and position-matched evaluation. Across MultiWOZ~2.4 and SGD, it consistently outperforms relevance-only and heuristic diversifiers, while remaining within practical latency constraints. 
Our analyses show that set-level diversity, not merely longer prompts or favorable exemplar order, is the main driver of JGA gains, offering a simple and actionable knob for deployment.

Limitations include dependence on training-split exemplars and specific diversity proxies, as well as a focus on English task-oriented benchmarks; formal optimality and robustness guarantees remain open. 
Future work will explore end-to-end learnable selectors with adaptive budgets, multilingual and open-domain settings, and online updating in production assistants, alongside tighter theoretical characterizations\citep{xiao2025diffusion}.

\bibliographystyle{ACM-Reference-Format}
\bibliography{refs}

\appendix

\section{Appendix}

\subsection{Additional technical details}

\paragraph{Default settings and practical notes.}
We use 
$m{=}0.2$ in \eqref{eq:metric_learning}, $\alpha{\in}[0.4,0.6]$, $\tau{\in}[0.3,0.5]$, $U{=}1$,
$L\in\{128,256\}$, $K\in\{4,6\}$, $\lambda_{\text{vec}}{\approx}0.6$, $\mu{\approx}0.05$, and $\tau_c{\in}[0.9,1.3]$.
For stability, we stratify dev evaluation by dialogue length and label frequency (micro/macro-F1) and report latency percentiles (P50/P90) of $t_{\text{tot}}$.
Finally, we cache history summaries for \emph{rolling} dialogues and reuse ANN warm probes across turns to amortize $t_{\text{ANN}}$.

\subsection{Implementation Details}
\label{sec:impl}

\subsubsection{LDRA memory construction.}
For each dataset, we build $\mathcal{M}=\{(x_i,y_i,e_i)\}$ from the \emph{training} split only. Exemplars store the user turn $x_i$ and the normalized state $y_i$. We canonicalize string values (lowercase, trim, normalize YES/NO/\texttt{not\_mentioned}) and map ontology synonyms to canonical forms used in the version’s official evaluation script.

\subsubsection{Encoders.}
The context/query encoder $f_\theta$ and the exemplar embedder $f_{\text{emb}}$ share weights. We start from an open-weight text encoder (base size) and fine-tune with the metric objective in Eq.~( \ref{eq:metric_learning} ) on training turns. Optimization: AdamW (lr=$2\!\times\!10^{-5}$, weight decay=$0.01$), batch size 256, 3 epochs, margin $m{=}0.2$, cosine learning-rate schedule with 500 warmup steps. Hard-negative mining is performed by periodic ANN sweeps over the memory.

\subsubsection{Retrieval and diversification.}
We index $e_i$ with an ANN library (HNSW/IVF). For the hybrid relevance in Eq.~(\ref{eq:hybrid}), we set $\lambda_{\text{vec}}{=}0.6$. We first take the top-$L$ candidates by $\operatorname{Rel}(\cdot)$, then run greedy selection with closed-form gains (Eqs.~(\ref{eq:gini_update})–(\ref{eq:text_update})) under constraints in Eq.~(\ref{eq:opt}). Unless otherwise specified:
$
L\in\{128,256\},\quad K\in\{4,6\},\quad
\alpha=0.5,\quad
\tau=0.4,\quad
U=1,\quad
\mu=0.05,\quad
\tau_c\in[0.9,1.3].
$
We tune $(\alpha,\tau,U,L,K,\lambda_{\text{vec}},\mu,\tau_c)$ on the dev split using JGA while enforcing a latency budget. 

The history summary $\sigma(C_{1:n-1})$ is truncated to at most 128 tokens. The exemplars are rendered as (user,domain/slots) pairs with minimal boilerplate.

\subsubsection{Prompting and verifier.}
We use the fixed instruction template of Eq.~(\ref{eq:pi}) for all LLMs. The verifier follows Eq.~(\ref{eq:score}) with YES/NO scoring and temperature calibration $\tau_c$. Generation parameters: temperature 0.2, top-$p$ 0.95, max output tokens 256; stop sequences constrain the model to the required JSON-like output.

\subsubsection{Hardware and runtimes.}
Training/fine-tuning and indexing run on 4$\times$A100\,40GB with 2$\times$Intel Xeon-class CPUs and 1\,TB RAM for fast ANN builds. Inference uses a single A100 or API endpoints for closed-source models. We report per-turn latency by component (\(t_{\text{ANN}}, t_{\text{DIV}}, t_{\text{PROMPT}}, t_{\text{LLM}}\)) and ensure the total meets an application budget of $\mathsf{B}\le 1.5$\,s; for voice-bot emulation we target $\mathsf{B}\le 800$\,ms with $L{=}128, K{=}4$.

\subsubsection{Input/Output Illustration}
\label{sec:io}
\textbf{Instruction (abridged).} \emph{``You are a customer-service agent. Given the dialogue history and the current user utterance, predict the active domain(s) and slot values. Output a JSON object.''} \\
\textbf{RAG exemplars (schematic).}
\texttt{(User)} \emph{``Can you help me get a taxi to Pizza Hut Fen Ditton?''} \(\Rightarrow\)
\texttt{(State)} \{\texttt{taxi: \{departure=..., destination=...\}}, \texttt{restaurant: \{name=..., price=expensive\}}\} \(\times K\). \\
\textbf{Current turn.}
\texttt{User:} \emph{``I would like a taxi from this College to THE HAPPY PARK.''} \\
\textbf{Model output (target).}
\texttt{taxi: \{departure="this College", destination="THE HAPPY PARK", arriveBy=not\_mentioned, leaveAt=not\_mentioned\}}.

\end{document}